\documentclass[10pt,twocolumn,letterpaper]{article}

\usepackage[pagenumbers]{cvpr} 

\usepackage{graphicx}
\usepackage{amsmath}
\usepackage{amssymb}
\usepackage{booktabs}
\usepackage{comment}
\usepackage{graphicx}
\usepackage{multirow}
\usepackage{xcolor}
\usepackage{paralist}
\usepackage{textcomp}

\usepackage{amsmath,amsfonts,bm}

\def\eg{\emph{e.g}\onedot} 
\def\ie{\emph{i.e}\onedot}

\def\wrt{w.r.t\onedot}

\def\eqref#1{equation~\ref{#1}}

\def\1{\bm{1}}

\def\eps{{\epsilon}}

\DeclareMathAlphabet{\mathsfit}{\encodingdefault}{\sfdefault}{m}{sl}
\SetMathAlphabet{\mathsfit}{bold}{\encodingdefault}{\sfdefault}{bx}{n}

\newcommand{\set}[1]{\ensuremath{\mathcal{#1}}}
\newcommand{\ingset}[0]{{\set{D}_{ing}}}

\newcommand{\func}[1]{\ensuremath{\mathrm{#1}}}

\newcommand{\subsdataset}[0]{Recipe1MSubs}

\newcommand{\GIN}[0]{GISMo}
\newcommand{\MLS}[0]{MLS}
\newcommand{\MLSFull}[0]{\textbf{M}ulti \textbf{L}ayer \textbf{S}equence}
\newcommand{\GPT}[0]{GPT2}
\newcommand{\GPTFull}[0]{GPT2-medium}
\newcommand{\GINFull}[0]{\textbf{G}raph-based \textbf{I}ngredient \textbf{S}ubstitution \textbf{Mo}dule}

\newcommand{\ISD}[0]{\func{ISD}}
\newcommand{\IE}[0]{\func{IE}}
\newcommand{\CE}[0]{\func{CE}}
\newcommand{\textapprox}{\raisebox{0.5ex}{\texttildelow}}
\usepackage[pagebackref,breaklinks,colorlinks]{hyperref}

\usepackage[capitalize]{cleveref}
\crefname{section}{Sec.}{Secs.}
\Crefname{section}{Section}{Sections}
\Crefname{table}{Table}{Tables}
\crefname{table}{Tab.}{Tabs.}

\def\cvprPaperID{1520} 
\def\confName{CVPR}
\def\confYear{2022}

\begin{document}

\title{Learning to Substitute Ingredients in Recipes}

\author{Bahare Fatemi$^{1}$\thanks{Work done during an internship at Meta AI (FAIR). Correspondence to: {\tt baharef@google.com}}\quad
Quentin Duval$^{2}$\quad
Rohit Girdhar$^{2}$\quad
Michal Drozdzal$^{2}$\quad
Adriana Romero-Soriano$^{2}$\\
$^{1}$ Google Research\quad
$^{2}$ Meta AI\\
}
\maketitle

\begin{abstract}
Recipe personalization through ingredient substitution has the potential to help people meet their dietary needs and preferences, avoid potential allergens, and ease culinary exploration in everyone's kitchen. To address ingredient substitution, we build a benchmark, composed of a dataset of substitution pairs with standardized splits, evaluation metrics, and baselines. We further introduce \GINFull{} (\GIN{}), a novel model that leverages the context of a recipe as well as generic ingredient relational information encoded within a graph to rank plausible substitutions. We show through comprehensive experimental validation that \GIN{} surpasses the best performing baseline by a large margin in terms of mean reciprocal rank. 
Finally, we highlight the benefits of \GIN{} by integrating it in an improved image-to-recipe generation pipeline, enabling recipe personalization through user intervention. Quantitative and qualitative results show the efficacy of our proposed system, paving the road towards truly personalized cooking and tasting experiences.\looseness-1
\end{abstract}

\section{Introduction}
\label{sec:intro}

In the current digital era, different food recipes from different cultures and regions in the world are spreading faster than ever. This widespread exposure to different types of cuisines inevitably sparks our interest to try new dishes. Being exposed to this wide variety of foods from different cuisines has widened our horizons and increased, in many cases, our curiosity and willingness to bring the exploration of new recipes home. However, cooking a new dish often requires some degree of recipe personalization due to the potential unavailability of certain ingredients, allergies (\eg to nuts or gluten), personal preferences, or lifestyle. Such recipe personalization usually occurs at the ingredient level, with some ingredients being substituted with alternatives. However, coming up with an ingredient substitution that would maintain the original dish taste as closely as possible is not trivial and requires broad cooking knowledge, as one ingredient might need to be substituted with different alternatives when appearing in different recipes. For example, \emph{applesauce} may be a good substitute for \emph{oil} when baking a cake but may not be as well suited when frying chicken. Overall, recipe personalization via ingredient substitution has the potential to improve our culinary experiences by helping us avoid ingredients that might be expensive, rare, allergic, or outside our dietary preferences.

\begin{figure}[t!]
   \centering
   \includegraphics[width=0.47\textwidth]{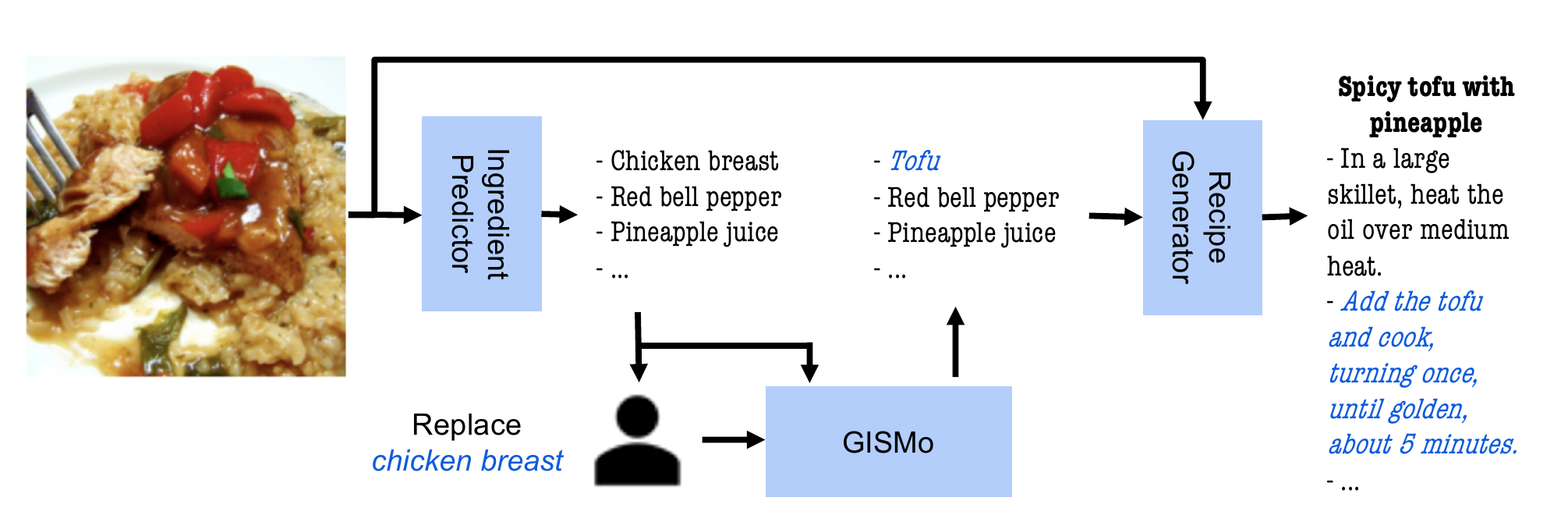}
   \caption{%
   \label{fig:introduction_noodles} %
    Overview of personalized image-to-recipe generation system with ingredient substitution. Given an ingredient to substitute, our ingredient substitution model, \GIN{}, suggests an alternative ingredient. The set of ingredients is updated and then used together with the image to generate the personalized recipe. 
    \vspace{-0.5cm}}
\end{figure}

Most existing approaches to ingredient substitution are based on information extracted from textual recipes, including simple statistics -- \eg Term Frequency–Inverse Document Frequency~\cite{shidochi2009finding,boscarino2014automatic,yamanishi2015alternative} --, and ingredient co-occurrences~\cite{achananuparp2016extracting,morales2021word,lawo2020veganaizer,kazama2018neural}. Only recently, researchers have started to explore advances in natural language processing (NLP) -- \eg word2vec~\cite{word2vec}, BERT~\cite{devlin2018bert}, and R-BERT~\cite{wu2019enriching} -- to obtain ingredient embeddings~\cite{foodbert,shirai2020identifying}, showing the benefits of representation learning to improve the task of ingredient substitution. This has been further emphasized with the development of node embedding methods to learn ingredient substitutions from the recently introduced large-scale FlavorGraph of ingredient relations and flavor molecules~\cite{park2021flavorgraph}. However, all the above-mentioned methods assume that ingredients have universal substitutions that are suitable in the context of any recipe. 
Moreover, obtaining high-quality ingredient substitutions is not sufficient to personalize recipes, since substituted ingredients oftentimes require adapting the cooking instructions. For example, when baking a cake, we may want to use \emph{banana} instead of \emph{egg}. However, the way to process eggs -- \emph{whisk eggs} -- is different from the way to process bananas -- \emph{mash bananas}. Therefore, learning-based approaches have been further extended in~\cite{li2021share} to not only propose ingredient substitutions but also alter the cooking instructions accordingly. This has been achieved by introducing a fully supervised pipeline that assumes access to pairs of original and personalized recipes, which are rather rare to find at scale and which are unfortunately not made publicly available.

Therefore, in this paper, we decouple the ingredient substitution from the recipe editing task, and instead, we focus on extracting ingredient substitution pairs from the flurry of readily accessible user comments found on recipe websites. As a result, we create a benchmark for ingredient substitutions called \subsdataset{}, with substitution pairs associated with the Recipe1M dataset~\cite{recipe1m} recipes. Our benchmark contains a dataset of valid ingredient substitution pairs together with their corresponding recipes as well as standardized dataset splits and metrics.

Moreover, we introduce \GINFull{} (\GIN{}), a novel ingredient substitution model which leverages not only the substitution context from a recipe but also generic ingredient relational information found in a large corpus of recipes, available through the FlavorGraph. More precisely, \GIN{} leverages recent advances in graph neural networks (GNNs) to capture common ingredient interactions within the learned ingredient embeddings, which are then updated with contextual information from a given recipe to predict plausible ingredient substitutions. We extensively evaluate our model and compare it with baselines, showing a performance improvement of \textapprox14\% in mean reciprocal rank over the best performing approach, and highlighting the value of leveraging complementary recipe-specific and generic ingredient relational information.

Finally, we plug \GIN{} into the inverse cooking system from~\cite{salvador2019inverse} to enable recipe editing. Inverse cooking first predicts the set of ingredients in an image and then generates a cooking recipe from the predicted ingredients and the input image. We modify the pipeline such that predicted ingredients are fed to \GIN{} together with a mask indicating the ingredient to be substituted, as specified by the user.
The resulting updated set of ingredients is used along the input image to generate the edited recipe, leveraging the compositionality captured by language models to operate outside of training distribution at inference time as opposed to requiring full supervision to generate personalized recipes. We further improve this pipeline by taking advantage of a vision transformer (ViT)~\cite{dosovitskiy2021an} backbone to predict the ingredients in the input image. Figure~\ref{fig:introduction_noodles} depicts the proposed personalized inverse cooking pipeline.  

The contributions of this paper can be summarized as:
\begin{compactitem}
    \item We create \subsdataset{}, a benchmark for ingredient substitution.
    \item We develop \GIN{}, a novel GNN-based ingredient substitution model, which leverages both recipe-specific context and generic ingredient relations to predict and rank plausible substitutes and improves upon previous approaches by a large margin.
    \item We extend the inverse cooking pipeline~\cite{salvador2019inverse} by enabling recipe personalization and show through extensive evaluation that the generated recipes effectively incorporate the modifications suggested by \GIN{}.
\end{compactitem}

\noindent Code for the personalized inverse cooking pipeline and the benchmark \subsdataset{} available at \url{
https://github.com/facebookresearch/gismo}.\looseness-1

\section{Ingredient substitution benchmark}
\label{sec:benchmark}

{\noindent \bf Ingredient substitution task.} The goal of ingredient substitution is to propose a valid substitution for a given ingredient in a recipe. Each recipe, $r$, is composed of a title, $t$, a set of ingredients, $\set{I}$, and a list of preparation instructions, $\textrm{P}$. During ingredient substitution, we modify the set of ingredients $\set{I}$ and update it into $\set{I'}$ with the substituted ingredient, assuming that an ingredient may be replaced by a single ingredient. 
More precisely, given an ingredient vocabulary $\ingset$ containing all ingredients in all recipes, the objective is to substitute a source ingredient $s \in \ingset$ for a target ingredient $y \in \ingset$ such that $\set{I'}=\set{I} \setminus \{s\} \cup \{y\}$. We call $(s, y)$ a \emph{substitution tuple}.  It is worth noting that substitution tuples are directional and may only be valid given the \emph{context} of the recipe. Therefore, we assume that we have access to a large dataset of $J$ recipes and substitution tuples $\tau = \{(s, y)_j, r_j\}_{j=1}^{j=J}$, and denote as $\tau_{train}$, $\tau_{val}$, and $\tau_{test}$ its respective training, validation and test splits.

{\noindent \bf Ingredient substitution dataset.} Recipe1M~\cite{recipe1m} is the largest publicly available collection of recipe data composed of $1,\!029,\!720$ recipes scraped from cooking websites. It contains $720,\!639$ training, $155,\!036$ validation and $154,\!045$ test recipes, with a title, a set of ingredients, a list of cooking instructions and (optionally) an image. However, this dataset does not contain ingredient substitution information. Inspired by~\cite{foodbert}, we crawled the Recipe1M recipe websites, identified all comments in the recipes, and extracted all sentences in the comments containing keywords potentially related to substitutions: ``instead'', ``substitute'', ``in place of'' and ``replace''. Next, we filtered 
the phrases containing substitution keywords as follows:
\begin{compactitem}
    \item We kept phrases that contain exactly two ingredients, as a different number of ingredients in the same sentence might not reflect a valid substitution.
    \item We kept phrases where the ingredients and substitution keywords are not \emph{far away} from each other to ensure keywords refer to the substitution (a distance of $ \geq 7$ is considered as \emph{far away}, value set by visual inspection).
    \item We kept phrases where \emph{only} one of the two identified ingredients is part of the original recipe ingredient set and store just the ingredient substitution pair by dropping all non-ingredients from the phrase.
\end{compactitem}
As a result, we obtain \subsdataset{} dataset that enables benchmarking models for the ingredient substitution problem. Each data point in \subsdataset{} contains a substitution tuple $(s, y)$, as well as the recipe ID (from the Recipe1M dataset) for which the substitution is valid. Using the recipe IDs and following the Recipe1M dataset ID splits, we get $49,\!044$, $10,\!729$ and $10,\!747$ substitution samples for training, validation, and test sets respectively. 

{\bf \noindent Evaluation metrics.} We follow the literature on recommender systems and link predictors on graphs~\cite{pinsage,transe,simple-plus} and evaluate the ingredient substitution task with Mean Reciprocal Rank (MRR) and Hit@$k$, $k \in \{1, 3, 10\}$. Both MRR and Hit@$k$ rely on \emph{ranking} the ingredients in $\ingset \setminus \{s\}$ for each recipe and source ingredient. The MRR computes the multiplicative inverse of the rank of the correct substitution answer, as determined by $y$. Hit@$k$ measures the proportion of correct substitutions that rank among the top $k$ answers. However, since there are cases for which different substitutions (target ingredients) might be proposed for the same source ingredient within a recipe, we avoid penalizing for such cases through their ranking by accepting all valid target ingredients as correct answers at evaluation time.

\section{\GIN{}}

\begin{figure}[t!]
   \centering
   \includegraphics[width=0.47\textwidth]{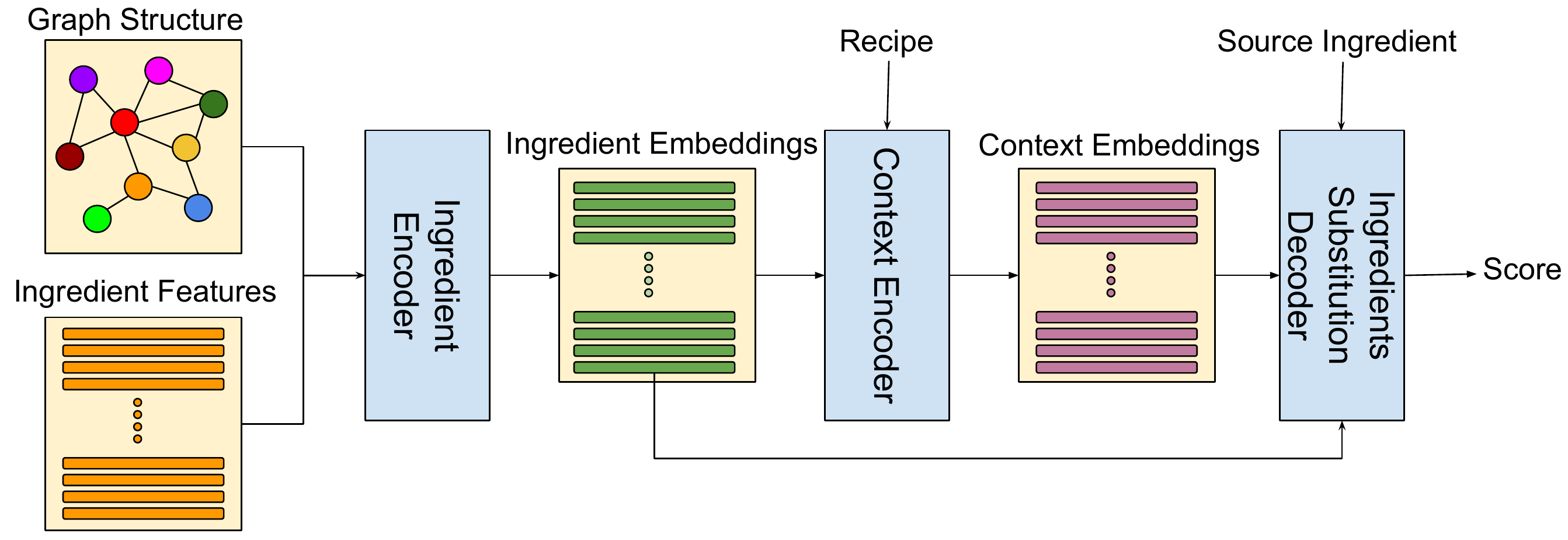}
   \caption{%
   \label{fig:gist} %
    Overview of \GIN{}, which is composed of: 1) ingredient encoder, 2) context encoder, and 3) ingredient substitution decoder. 
    The input graph models generic ingredient and food molecule-ingredient relations.}
\end{figure}
Given the context recipe $r$ and an ingredient to be substituted $s$, the objective of \GIN{} is to predict the target ingredient $y$ to use as a substitution.
\GIN{} consists of three components: 1) an ingredient encoder, 2) a context encoder, and 3) an ingredient substitution decoder. See Figure~\ref{fig:gist} for an illustration of the pipeline.  

{\bf \noindent Ingredient encoder (\IE{}).} This module takes as input a matrix of ingredient features as well as a graph structure -- represented using an adjacency matrix -- capturing ingredient relations and producing an embedding for each ingredient in the graph. 
The encoder is built from an optional embedding layer -- required when the ingredient features are one-hot encodings of the ingredients in $\ingset$ --, followed by multiple Graph Isomorphism Network (GIN)~\cite{gin} layers, given their properties, high performance, and ubiquitous use in the literature~\cite{hu2021open}. 
Each GIN layer is defined as follows:\looseness-1
\begin{equation*}\label{eq:GIN}
\mathbf{h}_v^{(l)} = f^{(l)}
\left ((1 + \eps^{(l)}) \mathbf{h}_v^{(l-1)} + \sum_{u \in \mathcal{N}(v)} ({e_{vu}} \mathbf{h}_u^{(l-1)}); \theta^{(l)}\right ),
\end{equation*}
where $\mathbf{h}_v^{(l)}$ represents the features of node $v$ at the $l$-th layer, $l \in [1, ..., L]$, $L$ is the number of GIN layers in the ingredient encoder, $f^{(l)}$ is a neural network function parameterized by $\theta^{(l)}$, $\epsilon^{(l)}$ is a learnable parameter, and $e_{vu}$ represents the weight of the edge connecting nodes $v$ and $u$ (a weight equal to 0 means no connection).

{\bf \noindent Context encoder (\CE{}).}
This module computes the context embeddings from a given recipe $r$. In \GIN{}, we capture the context by averaging the embeddings of the  ingredients in $\set{I}$. Additionally, in the ablations, we also consider the context captured in the recipe's title $t$ and use the pre-trained CLIP~\cite{clip} model to obtain the context embedding.

{\noindent \bf Ingredient substitution decoder (\ISD{}).}
This module assigns a score to all existing ingredients to determine the best substitution for a given source ingredient and recipe. More precisely, the substitution decoder takes as input the source ingredient, the ingredient embeddings from the ingredient encoder, and the context embedding from the context encoder. Then, it concatenates the ingredient embedding $\mathbf{h}_s^{(L)}$ corresponding to the source ingredient $s$ with the embedding $\mathbf{h}_v^{(L)}$ of every other ingredient $v$ as well as the context embedding $\mathbf{c}_r$ of a recipe $r$. The concatenation is fed through a multi-layer perceptron which predicts a score for each ingredient pair and recipe $\{(s, v), r\}$: $\phi_{s,v,r} = g \left (\mathbf{h}_s^{(L)} || \mathbf{h}_v^{(L)} ||  \mathbf{c}_r; \psi \right )$, where $g$ is a neural network function parameterized by $\psi$, and $||$ denotes concatenation. 

\begin{figure*}[ht!]
   \centering
   \includegraphics[width=0.95\textwidth]{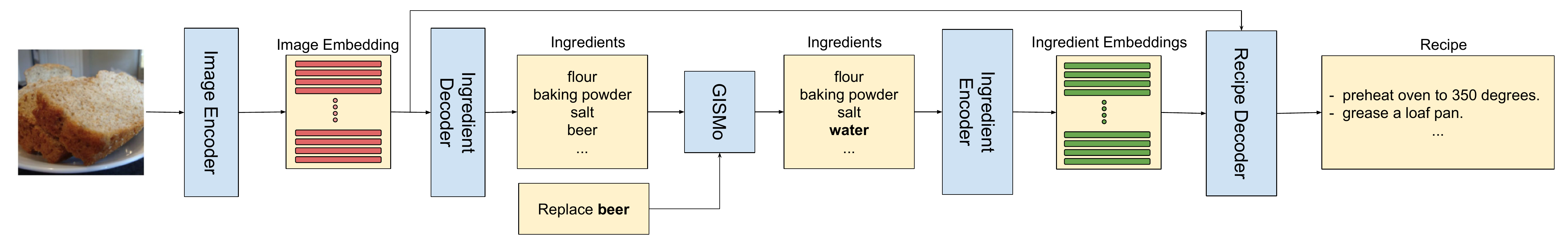}
   \caption{%
   \label{fig:inverse_cooking} %
    Overview of the personalized inverse cooking pipeline. An image encoder takes as input an image and produces an image embedding, which is fed to an ingredient decoder to predict the ingredients present in the image. The predicted ingredients are used as input to \GIN{} together with an ingredient to be substituted, specified by the user. \GIN{} outputs an updated set of ingredients, which is fed together with the image embedding to generate a personalized recipe.}
\end{figure*}

{\noindent \bf Objective function and training.} To learn the parameters of \GIN{} $\{\theta^{(1)}, ..., \theta^{(L)}\} \cup \{\epsilon^{(1)}, ..., \epsilon^{(L)}\} \cup \{\psi\}$, we use a self-supervised contrastive loss function~\cite{chen2020simple,baselines-strike}. The goal is to maximize the score $\phi_{s,y,r}$ for the substitution tuple $(s, y)$ and recipe $r$, and minimize the score $\phi_{s,v,r}$ for all other tuples $(s, v)$ given the same recipe $r$. For a data sample in $\tau_{train}$ (positive sample), we let $\mathcal{M}_{s,r} = \{(s, v_m), r)\}_{m=1}^{|\mathcal{M}|}$, such that $v_m \neq s$ and $v_m \neq y$,
be the corresponding set of negative substitution samples, and minimize the following objective: 
\begin{equation*}
\sum_{\{(s, y), r\} \in \tau_{train}}{-\log \frac{\mathrm{e}^{\phi_{s,y,r}}}{\displaystyle\sum_{v \in ~\{y\} ~\cup ~ \mathcal{M}_{s,r}}{\mathrm{e}^{\phi_{s,v,r}}}}}.
\end{equation*}
We train all three modules of \GIN{} end-to-end using mini-batch stochastic gradient descent~\cite{Bottou98on-linelearning}. At each training iteration, the model takes in a batch of substitution tuples and recipes. In practice, as we only have positive substitution samples available in our data, we generate negative samples for each positive sample following~\cite{transe,pinsage}, \ie for each positive sample, we obtain a subset of negative samples of cardinality $\zeta$ by replacing $y$ in the valid substitution tuples with random ingredients from $\ingset{}$, one at a time.

\section{Personalized inverse cooking pipeline}

To enable recipe personalization, we integrate \GIN{} in the inverse cooking pipeline described in \cite{salvador2019inverse} and introduce the \emph{personalized inverse cooking pipeline}. Figure~\ref{fig:inverse_cooking} illustrates the full pipeline at inference time. First, an image is fed to the image encoder and an image embedding is obtained. The image embedding is then used to predict a set of ingredients. \GIN{} module acts on the predicted ingredient set and substitutes one selected ingredient specified by the user. Next, the ingredient encoder of inverse cooking takes as an input the modified ingredient set and produces ingredient embeddings. Finally, both the ingredient embeddings of the modified ingredient set as well as the image embedding are fed into a recipe decoder that outputs a generated and personalized recipe. Important modifications \wrt \cite{salvador2019inverse} include the image encoder architecture, where we swap ResNet~\cite{HeZRS15} for Vision Transformers (ViT)~\cite{dosovitskiy2021an}. We further extend the ViT architecture by concatenating features corresponding to odd layers' outputs in the classification token as it proved to perform slightly better in our experiments than the vanilla ViT architecture. We refer to this modified ViT architecture as ViT \MLSFull{}, ViT MLS. Moreover, in contrast to~\cite{salvador2019inverse} and as suggested in~\cite{Pineda-1904} we train our image encoder using the binary cross-entropy loss. At training time, we remove the \GIN{} module and train the pipeline following the inverse cooking approach. Thus, at inference time, the personalized inverse cooking pipeline operates out of training distribution.

\section{Experiments}\label{sec:experiments}

In this section, we evaluate \GIN{} and the personalized inverse cooking pipeline. We start by providing experimental details, followed by results and ablations of both models.

\subsection{Experimental details}

{\bf \noindent Relations between ingredients.} To represent the relations between ingredients we use FlavorGraph~\cite{park2021flavorgraph}. FlavorGraph is a graph of ingredient relations extracted from Recipe1M~\cite{recipe1m}, which models ingredient co-occurrences in recipes. The graph also incorporates flavor molecules associated with the ingredients. In particular, it contains $6,\!653$ food ingredient nodes and $1,\!645$ flavor molecule (or chemical compound) nodes. Edges between two ingredient nodes reflect their association, computed as their normalized point-wise mutual information~\cite{bouma2009normalized}. Edges between ingredients and chemical compounds capture the chemical compounds in a food ingredient. Future work can study whether structure learning techniques~\cite{idgl,fatemi2021slaps} can identify latent structures between ingredients that help the task. 

{\bf \noindent Ingredient vocabulary.} Since FlavorGraph contains a subset of ingredients from Recipe1M, we process the ingredients in Recipe1M and \subsdataset{} such that they match those in FlavorGraph, \eg{} by merging singulars and plurals, different spellings of the same ingredient, or very infrequent ingredients with their closest match. As a result, our ingredient vocabulary contains $6,\!653$ ingredients. Note that the original inverse cooking pipeline~\cite{salvador2019inverse} uses a pre-processed vocabulary of $1,\!488$ ingredients, which is more than $4\times$ smaller than the FlavorGraph-based one.

{\noindent \bf \GIN{} details.} 
We implement our model in PyTorch \cite{paszke2017automatic}, and use deep graph library (DGL) \cite{wang2019deep} for the sparse operations. We use the Adam~\cite{adam} optimizer, and perform early stopping and hyperparameter tuning based on the MRR on the validation set in all cases. 
We defer more details to the supplementary material.

{\noindent \bf Personalized inverse cooking pipeline details.} We train our pipeline following~\cite{salvador2019inverse}. We start by training the image-to-ingredient prediction module, and then train the recipe generator using teacher forcing both from the ground truth ingredients and recipe. We add weight decay in the recipe generation training stage, as we found it helped mitigate the overfitting caused by the increased number of ingredients in our vocabulary. The weight decay was set to $10^{-5}$ after a sweep among $[0.0, 10^{-6}, 10^{-5}, 10^{-4}, 10^{-3}]$. Note that \GIN{} is trained on its own and only used at inference time within the pipeline.\looseness-1 

\begin{table}[t]
\begin{center}
\resizebox{0.49\textwidth}{!}{%
\setlength{\tabcolsep}{2pt}
\begin{tabular}{lcccc}
 \textbf{Model}  & \textbf{MRR} & \textbf{Hit@1} & \textbf{Hit@3} & \textbf{Hit@10}\\\midrule
 Random & $0.15 \pm 0.02$ & $0.02 \pm 0.01$ & $0.05 \pm 0.03$ & $0.17 \pm 0.05$\\
 Mode & $1.62 \pm 0.00$ & $1.61 \pm 0.00$ & $1.61 \pm 0.00$ & $1.61 \pm 0.00$\\
 LT &  $2.53 \pm 0.00$ & $0.15 \pm 0.00$ & $0.88 \pm 0.00$ & $5.80 \pm 0.00$\\
 Freq &  $3.81 \pm 0.00$ & $1.61 \pm 0.00$ & $2.77 \pm 0.00$ & $6.41 \pm 0.00$\\
 LT+Freq & $27.66 \pm 0.00$ & $18.10 \pm 0.00$ & $32.26 \pm 0.00$ & $47.50 \pm 0.00$\\
 \midrule
 Food2Vec & $9.65 \pm 0.00$ & $4.79 \pm 0.00$ & $10.71 \pm 0.00$ & $19.99 \pm 0.00$\\
 FoodBERT & $9.32 \pm 0.01$ & $4.53 \pm 0.01$ & $9.27 \pm 0.02$ & $22.79 \pm 0.03$\\
 R-FoodBERT & $10.09 \pm 0.01$ & $7.13 \pm 0.01$ & $12.48 \pm 0.02$ & $15.31 \pm 0.02$\\
 \midrule 
 Metapath2vec &  $0.79 \pm 0.00$ & $0.03 \pm 0.00$ & $0.03 \pm 0.00$ & $2.18 \pm 0.00$\\
 PinSAGE & $17.86 \pm 0.51$ & $6.93 \pm 0.61$ & $20.59 \pm 0.49$ & $41.84 \pm 1.20$\\
 \midrule
\GIN{} (Ours) &  $\bf{31.51} \pm 0.10$ & $\bf{20.56} \pm 0.14$ & $\bf{35.89} \pm 0.16$ & $\bf{54.27} \pm 0.19$\\
\end{tabular}
}
\caption{Test results on \subsdataset{}. Our method outperforms the baselines on all metrics. Results are averaged over 5 runs, and the mean $\pm$ std are reported.\vspace{-0.5cm}}
\label{table:ingredient_substitution}
\end{center}
\end{table}

\subsection{Evaluating \GIN{}}

{\bf \noindent Baselines.} 
We consider the following statistical, language model-based, and graph-based baselines. \underline{Random}: This baseline randomly ranks all possible ingredient substitutions given a source ingredient and a recipe. \underline{Mode}:
This baseline finds the ingredient that appears the most often as a target ingredient across all data samples in $\tau_{train}$, and uses it as a proposed substitution in all cases. \underline{Lookup table (LT)}: For each source ingredient, this baseline assigns rank 1 to all target ingredients that appear together with said source ingredient in the substitution tuples of $\tau_{train}$.
\underline{Frequency (Freq)}: This baseline counts the number of times each $v \in \ingset$ is suggested as a target ingredient in $\tau_{train}$, and uses these frequencies to rank ingredient substitutions. \underline{Lookup table + Frequency (LT+freq)}: This baseline combines the LT and Freq ranking approaches and computes how often a target ingredient appears for each source ingredient in $\tau_{train}$. Those frequencies are then used to rank ingredient substitutions for each source ingredient. \underline{Food2Vec, FoodBERT, and R-FoodBERT}: Those are language model-based approaches introduced in~\cite{foodbert}. Food2Vec and FoodBERT finetune respectively word2vec~\cite{word2vec} and BERT~\cite{devlin2018bert} models on cooking instructions of the Recipe1M dataset, and then use the updated word embeddings to suggest ingredients substitutions. The assumption is that ingredients that are close to each other in the embedding space are good candidate substitutions for each other. R-FoodBERT further finetunes the FoodBERT model on user comments crawled from the Recipe1M recipe websites, which contain substitution information. \underline{Metapath2vec~\cite{park2021flavorgraph}}: This baseline implicitly learns structural node embeddings by leveraging the metapath2vec~\cite{metapath2vec}, an extension of node2vec~\cite{grover2016node2vec} that operates on graphs with different node types. The approach enhances the node embedding learning of metapath2vec by including chemical property information.
\looseness-1

{\noindent \bf Results.} 
Results are averaged over 5 runs and are reported in Table~\ref{table:ingredient_substitution}. The strongest statistical approach is LT+Freq, which is a combination of LT and Freq baselines, and achieves an MRR score of $27.66$. We can also see that using LT and Freq on their own leads in both cases to relatively poor MRR scores. Among the language model baselines, R-FoodBERT performs the best, reaching $10.09$ MRR score. However, none of the language model baselines outperforms the LT+Freq baseline, suggesting that direct access to ingredient substitution tuples is beneficial. When it comes to node embeddings learned from random walks on the graph (metapath2vec), results drop by a large margin and are among the weakest ones. This is perhaps unsurprising given that FlavorGraph captures ingredient co-occurrences but has no direct information about plausible ingredient substitutions. 
To assess the importance of leveraging the information captured by FlavorGraph in a more explicit way and in the context of ingredient substitution, we borrow a model from the GNN-based recommender system literature and simplify it to fit our task. The simplification comes from the absence of user-item interactions in our data. The model~\cite{pinsage}, PinSAGE, leverages GraphSAGE~\cite{NIPS2017_Hamilton} layers to learn node embeddings through a max-margin ranking loss. The learned embeddings are then used to suggest ingredient substitutions through nearest neighbor lookups. As shown in the table, PinSAGE outperforms all the language model-based approaches, with an MRR of $17.86$, suggesting that the generic ingredient relational information captured by the FlavorGraph is useful\footnote{We also tested a variant of PinSAGE with GIN layers and observed that the results were unaffected.}. However, the approach still falls far behind the LT+Freq baseline. Finally, we notice that our model can outperform all alternative approaches, improving the MRR score by $14\%$ \wrt the LT+Freq baseline. Similar trends hold for the Hit@ metrics, with \GIN{} consistently outperforming other methods.

\begin{table}[t]
\begin{center}
\resizebox{0.49\textwidth}{!}{
\setlength{\tabcolsep}{3pt}
\begin{tabular}{lccc}
\textbf{Model} & ID & OOD & Average\\\midrule
Random  & $0.15 \pm 0.03$ & $0.13 \pm 0.01$ & $0.14 \pm 0.01$\\
Mode & $2.31 \pm 0.00$ & $0.39 \pm 0.00$ & $1.35 \pm 0.00$\\
LT & $3.91 \pm 0.00$ & $0.02 \pm 0.00$ & $1.97 \pm 0.00$\\
Freq & $5.11 \pm 0.00$ & $1.44 \pm 0.00$ & $3.27 \pm 0.00$\\
LT+Freq & $42.95 \pm 0.00$ & $0.03 \pm 0.00$ & $21.49 \pm 0.00$\\ \midrule
Food2Vec & $11.93 \pm 0.00$ & $4.74 \pm 0.00$ & $8.33 \pm 0.00$\\
FoodBERT & $11.69 \pm 0.01$ & $4.48 \pm 0.05$ & $8.08 \pm 0.03$\\
R-FoodBERT & $13.06 \pm 0.01$ & $3.76 \pm 0.03$ & $8.41 \pm 0.01$\\
 \midrule
 Metapath2vec & $0.91 \pm 0.0$ & $0.61 \pm 0.0$ & $0.76 \pm 0.0$ \\ 
PinSAGE & $25.09 \pm 0.49$ & $4.75 \pm 0.43$ & $14.92 \pm 0.46$\\ 
 \midrule
\GIN{} (Ours) & $\bf{46.53} \pm 0.17$ & $\bf{5.77} \pm 0.31$ & $\bf{26.15} \pm 0.22$\\
\end{tabular}
}
\caption{Stratified in distribution (ID) and out of training distribution (OOD) MRR test results.}\vspace{-0.5cm}
\label{table:stratified_results}
\end{center}
\end{table}

{\bf \noindent Stratified results for ingredient substitution.}
To gain further understanding of models' behavior, we report the MRR score on the stratified dataset in Table~\ref{table:stratified_results}. In particular, we investigate how well the model generalizes if the substitution tuple appears in the training set but with a different context -- referred to as in distribution scenario (ID) --, and contrast it with a situation when the substitution tuple does not appear in the training set -- referred to as out of training data distribution scenario (OOD).
The results show that \GIN{} outperforms all baselines consistently both ID and OOD. Since the previously discussed overall results are dominated by the ID case, we focus the discussion on the OOD scenario. Leveraging the graph structure of FlavorGraph enables node ingredients in OOD tuples to update their embeddings through messages from their neighbors in the GNN forward pass. Therefore, the message passing information helps \GIN{} to be able to handle OOD samples better than the baselines. The stratified results of PinSAGE further confirm this hypothesis, achieving OOD results comparable to those of \GIN{} even though its performance ID suffers.



\subsection{Ablative analysis} 

{\noindent \bf Importance of different modules.}
We ablate the importance of different modules in the architecture by, 1) removing the ingredient encoder (\IE{}) to assess the importance of leveraging the graph structure, and 2) the context encoder (\CE{}) to assess the importance of leveraging the recipe context. We also measure the contribution of different data modalities used in the context encoder (ingredient set vs recipe title). The ablation results are reported on the validation set and depicted in Table~\ref{table:modalities}. 

As it can be seen, the best performing version of \GIN{} has the three modules of \IE{}, \CE{}, and \ISD{}, suggesting that both contextual recipe information, as well as ingredient relations, are important for the ingredient substitution task. The \ISD{}-based model, which takes as input the ingredient features directly and does not leverage context nor the graph structure, achieves the lowest performance in terms of MRR. When adding context information to \ISD{}, improvements are dependent on the context used, with recipe titles appearing to be beneficial (\CE{}$_t$ + \ISD{}). When considering the graph structure (\IE{} + \ISD{}), we also notice improvements, which are however below those achieved when leveraging context and graph structure at the same time. When considering both the graph structure and contextual information (\IE{} + \CE{} + \ISD{}), the kind of contextual information used becomes less important, with all variants performing similarly. However, it is worth noting that \IE{} + \CE{}$_\set{I}$ + \ISD{} requires less computation at inference time since it does not need any extra module to embed the recipe title. 

\begin{table}[t]
\begin{center}
\setlength{\tabcolsep}{2pt}
\resizebox{0.49\textwidth}{!}{%
\begin{tabular}{lcccc}
\textbf{Context} & \textbf{MRR} & \textbf{Hit@1} & \textbf{Hit@3} & \textbf{Hit@10}\\\midrule
\ISD{} & $29.33 \pm 0.09$ & $18.93 \pm 0.16$ & $33.72 \pm 0.15$ & $50.89 \pm 0.11$\\ \midrule
\IE{} + \ISD{} & $29.69 \pm 0.08$ & $18.97 \pm 0.12$ & $33.93 \pm 0.08$ & $51.82 \pm 0.17$\\ \midrule
\CE{}$_\set{I}$ + \ISD{} & $28.97 \pm 0.14$ & $18.73 \pm 0.19$ & $32.71 \pm 0.12$ & $50.10 \pm 0.39$\\
\CE{}$_t$ + \ISD{} & $29.95 \pm 0.08$ & $20.12 \pm 0.11$ & $34.24 \pm 0.19$ & $50.04 \pm 0.31$\\
\CE{}$_{\set{I},t}$ + \ISD{} & $29.04 \pm 0.32$ & $18.85 \pm 0.31$ & $32.80 \pm 0.33$ & $49.98 \pm 0.27$\\
\midrule
\IE{} + \CE{}$_\set{I}$ + \ISD{} & $\bf{31.25} \pm 0.12$ & $20.52 \pm 0.16$ & $\bf{35.32} \pm 0.23$ & $\bf{53.37} \pm 0.16$\\
\IE{} + \CE{}$_t$ + \ISD{} & $\bf{31.33} \pm 0.11$ & $\bf{20.72} \pm 0.18$ & $\bf{35.56} \pm 0.21$ & $53.09 \pm 0.11$\\
\IE{} + \CE{}$_{\set{I},t}$+ \ISD{} & $\bf{31.37} \pm 0.07$ & $\bf{20.78} \pm 0.10$ & $\bf{35.59} \pm 0.26$ & $53.05 \pm 0.11$ \\
\end{tabular}
}
\caption{Ablations to assess the importance of leveraging context and generic ingredient relations in \GIN{}. Results are reported on the validation set.}
\label{table:modalities}
\end{center}
\end{table}

{\noindent \bf Exploring different ingredient features as input.}
Here, we 
test the effect of ingredient feature initialization on the overall model performance. Since the FlavorGraph does not come with ingredient features, \GIN{} uses a randomly initialized embedding layer which is updated during training. In this ablation, we leverage \emph{pretrained} ingredient features learned by language models or through node structural encodings instead. In particular, we consider metapath2vec augmented with chemical structures~\cite{park2021flavorgraph}, random walks (node2vec)~\cite{grover2016node2vec}, as well as FoodBERT ingredient features~\cite{foodbert}. Additionally, we take the pretrained ingredient features which best perform OOD (FoodBERT) and enable finetuning of those features at training time~\footnote{We use L2 regularization on the learnable values to avoid the input features going too far from the FoodBERT embeddings}. From the results, we observe that leveraging pretrained ingredient features leads to an overall MRR performance that is comparable to the one achieved by learnable randomly initialized embeddings. However, when it comes to OOD results, structural node embeddings appear to bring some performance improvements, with node2vec achieving better results than metapath2vec. Those improvements are even more remarkable when leveraging embeddings from FoodBERT. However, those trends are not preserved ID. Finally, when enabling gradient propagation through FoodBERT, the benefits of generalizing OOD are reduced.

\begin{table}[t]
\begin{center}
\resizebox{0.49\textwidth}{!}{
\setlength{\tabcolsep}{3pt}
\begin{tabular}{lcccc}
\textbf{Input features} & ID & OOD & Avg & \textbf{Overall}\\
\cmidrule(lr){1-1}\cmidrule(lr){2-4}\cmidrule(lr){5-5}
Random init. & $46.65 \pm 0.14$ & $5.69 \pm 0.43$ & $26.17 \pm 0.24$ & $31.25 \pm 0.12$\\
Metapath2vec \cite{park2021flavorgraph} & $46.29 \pm 0.16$ & $5.19 \pm 0.51$ & $25.74 \pm 0.31$ & $31.22 \pm 0.11$\\
Node2vec \cite{grover2016node2vec} & $46.26 \pm 0.08$ & $6.52 \pm 0.51$ & $26.39 \pm 0.27$ & $31.30 \pm 0.08$\\
FoodBERT \cite{foodbert} & $44.37 \pm 0.17$ & $8.44 \pm 0.53$ & $26.40 \pm 0.28$ & $30.35 \pm 0.12$\\
FoodBERT finetuned & $45.88 \pm 0.35$ & $8.02 \pm 0.28$ & $26.95 \pm 0.16$ & $31.18 \pm 0.21$\\
\end{tabular}
}
\caption{Ablation on input features. Stratified MRR results for \GIN{} with different input features on the validation set.\vspace{-0.5cm}}
\label{table:input_features}
\end{center}
\end{table}

{\noindent \bf Are substitutions directional?}
In this experiment, we test if we can swap the source and target ingredients during training and maintain or improve the system's performance. To this end, at training time, we introduce a data augmentation layer that swaps the source and target ingredients 
randomly with a probability of $0.5$. The validation set MRR for this experiment reaches $30.61 \pm 0.10$, while the original \GIN{} model obtained $31.25 \pm 0.12$, thus we observe a decrease of $0.64$ on average. This suggests that the directionality of the ingredients in the substitution tuple provides an important signal. This drop can be further explained by looking at examples, \eg when cooking a cake one might substitute sugar for applesauce to make the cake healthier; however, it would be unlikely to substitute applesauce for sugar. In general, we observe that users tend to substitute rare, allergen, expensive, or unhealthy ingredients for their frequent, non-allergen, cheaper, or healthier versions.


\subsection{Personalized inverse cooking evaluation}

We start by evaluating the modifications we introduced to the inverse cooking pipeline, namely the architecture of the ingredient encoder. Table \ref{table:inverse_cooking_architectures} compares the performance of the ResNet50-based inverse cooking pipeline of \cite{salvador2019inverse} to the ones using ViT and ViT \MLS{} image encoders on the Recipe1M dataset. Note that in this experiment, we do not use \GIN{} module. The table reports the F1 score of the image-to-ingredients prediction task as well as the perplexity (PPL) of the generated recipes when using either the predicted ingredients or the ground truth ingredients. We observe that the proposed ViT \MLS{} obtains the best results on average, reaching an F1 score of $33.2\%$ and a PPL from predicted ingredients of $9.47$, improving the ResNet50 baseline by $2.9$ and $0.22$, respectively.\looseness-1

\begin{table}[t]
\begin{center}
\resizebox{0.48\textwidth}{!}{
\begin{tabular}{ l  c c c } 
  \textbf{Model} & \textbf{F1} $\uparrow$ & \textbf{GT ingr PPL} $\downarrow$ & \textbf{Pred Ingr PPL} $\downarrow$ \\ 
 \midrule
  ResNet50~\cite{salvador2019inverse} & $30.3 \pm 0.03$ & $7.35 \pm 0.03$ & $9.69 \pm 0.03$ \\ \midrule
  ViT-32 & $31.5 \pm 0.13$ & $\mathbf{7.18 \pm 0.02}$ & $9.80 \pm 0.04$ \\ 
  ViT-16 & $\mathbf{33.1 \pm 0.12}$ & $\mathbf{7.19 \pm 0.02}$ & $9.57 \pm 0.04$ \\ 
  \midrule
 ViT-32 \MLS{} & $31.6 \pm 0.10$ & $\mathbf{7.18 \pm 0.01}$ & $9.67 \pm 0.03$ \\ 
 ViT-16 \MLS{} & $\mathbf{33.2 \pm 0.04}$ & $\mathbf{7.17 \pm 0.01}$ & $\mathbf{9.47 \pm 0.04}$ \\
\end{tabular}}
\caption{Image encoder comparisons in terms of F1 score and PPL from both ground truth and predicted ingredients. Results report mean $\pm$ std over 5 runs with different seeds. ViT MLS achieves the best performance on average across all metrics.\vspace{-0.5cm} }
\label{table:inverse_cooking_architectures}
\end{center}
\end{table}


\begin{table}[t]
\begin{center}
\resizebox{0.49\textwidth}{!}{
\begin{tabular}{ l c c c c}
  \\
  \textbf{Metric} & \textbf{No sub.} & \textbf{GT sub.} & \textbf{\GIN{} sub.} & \textbf{Rand. sub.}
  \\ 
 \cmidrule(lr){1-1}\cmidrule(lr){2-5}
 PPL  &  $6.60 \pm 0.01$ & $6.84 \pm 0.01$ & $6.84 \pm 0.01$ & $6.91 \pm 0.01$
 \\
 \GPT{} PPL         & $8.58 \pm 0.31$ & $8.50 \pm 0.25$ & $8.51 \pm 0.30$ & $8.35 \pm 0.28$
 \\ 
 Recipe length         & $75.42 \pm 1.32$ & $75.53 \pm 1.34$ & $75.39 \pm 1.47$ & $76.37 \pm 1.42$
 \\ 
 Unique words   & $41.99 \pm 0.67$ & $41.96 \pm 0.62$ & $41.95 \pm 0.67$ & $42.35 \pm 0.66$
 \\ 
 IoU      & $48.02 \pm 0.73$ & $47.19 \pm 0.74$ & $47.87 \pm 0.88$ & $44.01 \pm 0.70$
 \\ 
\end{tabular}}
\caption{Measuring the effect of ingredient substitutions on the quality of the generated recipes. No sub.: no substitution, GT sub.: using ground truth substitutions from \subsdataset{}, \GIN{} sub.: using predicted substitutions, Rand. sub.: using random substitutions.\vspace{-0.5cm}
}
\label{table:subs_metrics}
\end{center}
\end{table}

\begin{figure*}[ht!]
   \centering
   \includegraphics[width=0.98\textwidth]{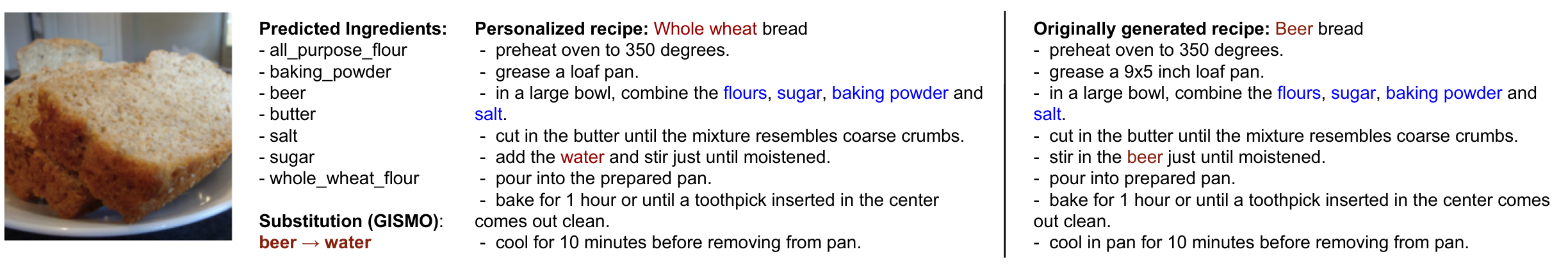}
   \caption{%
   \label{fig:inverse_cooking_qualitative} %
    Example of a personalized recipe. The image of a ``beer bread loaf'' is fed to the inverse cooking pipeline. We then ask to substitute \emph{beer} and \GIN{} suggests using \emph{water} instead. The generated recipe is entitled by the model as ``whole wheat bread''. When comparing the personalized (left) recipe with the non-personalized one (right), we observe that only one instruction is modified (fifth from the top). \vspace{-0.5cm}
    }
\end{figure*}


Next, we evaluate the effect of \GIN{} on the quality of the generated recipes. As we do not have ground truth recipes with the modified ingredient set, we rely on multiple complementary metrics to judge the quality of generated recipes. First, we consider the perplexity of personalized recipes from modified ingredients \wrt the ground truth recipe (PPL). We compare the personalized recipes, both from ground truth substitutions and predicted substitutions, to the ones generated from the original ground truth ingredients to determine how much the personalized recipes deviate from the non-personalized ones. Second, we compute the perplexity of the generated recipes \wrt a pretrained \GPTFull{}~\cite{radford2019language} (\GPT{} PPL) to further assess the quality of the generated recipes, following
~\cite{Zhang*2020BERTScore:,BLEURT_Sellam}. %
The \GPT{} PPL metric is obtained by concatenating the generated recipes to an initial prompt "recipe instructions:" which serves as a context for \GPT{} to switch to a \emph{recipe world}. 
As \GPT{} has been shown to be sensitive to text features such as text length~\cite{miaschi-etal-2021-makes}, we complement the metric by reporting the average length of the generated recipes, and the number of unique vocabulary words in them. These metrics help us identify potential short and non-specific generations such as ``mix all ingredients''. Finally, we include the intersection over union (IoU) to measure the consistency between the ingredients used as a conditioning and those appearing in the generated recipes. However, since the recipes often contain ambiguous references to the ingredients such as \emph{pepper} to refer to \emph{cayenne pepper} or \emph{dry ingredients} to refer to a bunch of dry ingredients all together, the results of this metric should be interpreted as a rough indication of consistency. We complete our analysis with a qualitative evaluation of the generated recipe by visual inspections. 

The results for this experiment are reported in Table \ref{table:subs_metrics}, which shows average metrics for 25 models obtained by changing random seed (5 runs of the \GIN{} model and 5 runs of the inverse cooking one). The table measures the effect of ingredient substitutions on the quality of the generated recipes. In particular, it contrasts the effect of introducing no substitutions, with substitutions coming from the ground truth tuples of \subsdataset{}, the substitutions predicted by \GIN{}, as well as random substitutions.
Based on the results, we can conclude that
the inverse cooking pipeline does not deviate too much from the generated recipe when we substitute the ground truth ingredients and maintains ingredient consistency in the generated recipes as indicated by the IoU metric. Recipes generated with substitutions do not increase the number of non-specific recipes -- \ie, the ingredient IoU stays within standard deviation for all three tested scenarios and the recipe length is also approximately maintained, suggesting that leveraging the compositionality captured by language models is a good alternative to collecting supervised recipe pairs when it comes to recipe editing. When it comes to random substitution, we observe an expected increase in PPL and a decrease in IoU \wrt other substitution strategies. However, the \GPT{} PPL does not appear to be affected (within std).

Finally, we report a qualitative result in Figure~\ref{fig:inverse_cooking_qualitative}. One can observe that the \GIN{} module proposes a reasonable substitution in this case -- suggesting to use \emph{water} instead of \emph{beer} --, and that the generated recipe reflects well the changes in the personalized recipe. When comparing the personalized recipe to the recipe obtained by the inverse cooking pipeline, we observe that the substitution proposed by \GIN{} affected only one cooking instruction and modified the recipe title, changing it from \emph{beer bread} to \emph{whole wheat bread}.

\section{Related work}\label{sec:related_work}

{\noindent \bf Food understanding.} Food understanding has been gaining momentum in the computer vision community since the introduction of
large-scale datasets~\cite{bossard2014food}. The researchers have investigated a variety of food-related recognition tasks, such as image classification~\cite{horiguchi2018personalized,lee2018cleannet}, ingredient classification~\cite{chen2017cross}, calorie estimation~\cite{Meyers_2015_ICCV} quantity prediction~\cite{chen2012automatic}, and food question answering~\cite{FoodQA}. More recently, the inception of Recipe1M dataset~\cite{marin2019recipe1m+} triggered the research in cross-modal recipe retrieval~\cite{Retrieval_Lien,Zhu_2019_CVPR,salvador2021revamping}, image-to-recipe generation~\cite{salvador2019inverse,SAGN_Wang,nishimura-etal-2019-procedural,chandu-etal-2019-storyboarding} as well as recipe-to-image generation~\cite{ChefGAN,CookGANZhu_2020_CVPR,CookGANHan}. It is worth noting that topics of food understanding have also been explored in the natural language processing literature, where significant effort has been devoted to advancing recipe generation~\cite{NLP_Pan,NLP_singh}, and food recommendation systems~\cite{jiang2019market2dish}.

{\noindent \bf Ingredient substitution and recipe editing.}
The vast amount of cooking recipes available online has triggered the exploration of ingredient substitution methods which leverage textual information. In particular,~\cite{shidochi2009finding,boscarino2014automatic,yamanishi2015alternative} have applied statistical methods such as Term Frequency–Inverse Document Frequency on recipe instructions to propose valid ingredient substitution. Similarly, recipe editing has been studied by leveraging ingredient co-occurrences~\cite{achananuparp2016extracting,morales2021word,shirai2020identifying,lawo2020veganaizer,kazama2018neural} to meet dietary restrictions and adapt recipes to regional cuisines. More recently, researchers have leveraged recent advances in the natural language processing literature -- \eg, word2vec~\cite{word2vec}, BERT~\cite{devlin2018bert}, and R-BERT~\cite{wu2019enriching} -- to learn ingredient embeddings from recipe instructions and/or user comments~\cite{foodbert}. The learned embeddings are then used in the context of ingredient substitution by applying the $k$-nearest neighbors algorithm~\cite{fix1951discriminatory}. Moreover, \cite{li2021share} has tackled the problem of recipe editing by collecting pairs of original and edited recipes. Finally, the introduction of a large-scale graph of ingredient co-occurrences and food molecules, the FlavorGraph~\cite{park2021flavorgraph}, has triggered the exploration of node embedding methods. In particular, metapath2vec~\cite{metapath2vec,park2021flavorgraph} has been extended to incorporate chemical properties, and as a result, improve food pairing.

\section{Discussion}

{\noindent \bf Conclusions.} In this paper, we tackled the problem of recipe personalization by decoupling ingredient substitution from recipe editing. We addressed ingredient substitution by introducing a benchmark for the task. We then devised a graph-based ingredient substitution module able to leverage recipe-specific context and generic ingredient relational information encoded as a graph simultaneously to predict plausible substitutions. Finally, we took advantage of the compositionality captured by pretrained language models to integrate the proposed ingredient substitution model within the inverse cooking pipeline, and highlighted the effectiveness of the system to come up with plausible substitutions and adequate recipe modifications. Our research takes one step forward to increase creativity in everyone's kitchen through personalized tasting experiences.

{\noindent \bf Limitations.} We consider a single ingredient substitution. Although this setup works well in our experiments, the users might want to substitute multiple ingredients at once. Moreover, by design we always consider one-to-one substitutions, meaning that a single source ingredient can be substituted by one target ingredient. This might be unrealistic in some scenarios, where a single ingredient should be substituted by multiple ingredients. 
Future work can study whether learning with hypergraphs (\eg{} ~\cite{feng2019hypergraph,knowledge-hypergraphs}) can help with many-to-many substitutions.

{\noindent \bf Potential negative societal impact.} Our work is based on the Recipe1M dataset that was scraped from mostly US-centric food webpages. Thus, as a result, the learned substitutions might have a bias toward western cuisines, and the system might not work equally well for everyone.\looseness-1

{\small
\bibliographystyle{ieee_fullname}
\bibliography{egbib}
}

\appendix

\section{\GIN{} Implementation Details}
We implemented our model in PyTorch \cite{paszke2017automatic}, used deep graph library (DGL) \cite{wang2019deep} for the sparse operations, and used Adam~\cite{adam} as the optimizer. We performed early stopping and hyperparameter tuning based on the MRR on the validation set for all models, and used a maximum number of epochs of $1000$. We tuned the learning rate $lr$ using the set $\{0.005, 0.001, 0.0005, 0.0001, 0.00005, 0.00001, 0.000005\}$, and both the number of features $F$ and ingredient embedding dimension $d$ using the set $\{200, 300, 400, 500, 600\}$. 

Our final model consists of an ingredient encoder (\IE{}) composed of two GIN~\cite{gin} layers, and an ingredient substitution decoder (\ISD{}) composed of a multi-layer perceptron with three layers. The model uses $F=300$. The input ingredient features are one-hot encodings and the ingredient embedding dimension is set to $d=300$. The model is trained with a learning rate $lr=0.00005$, and is regularized using a weight decay of $0.0001$ and a dropout rate of $0.25$.\looseness-1

\section{Visual inspection of recipes}

All ingredient predictions and recipe generations visualized in the figures are based on our best ViT-16 \MLS{}.

\begin{figure*}[b]
   \includegraphics[width=0.89\textwidth]{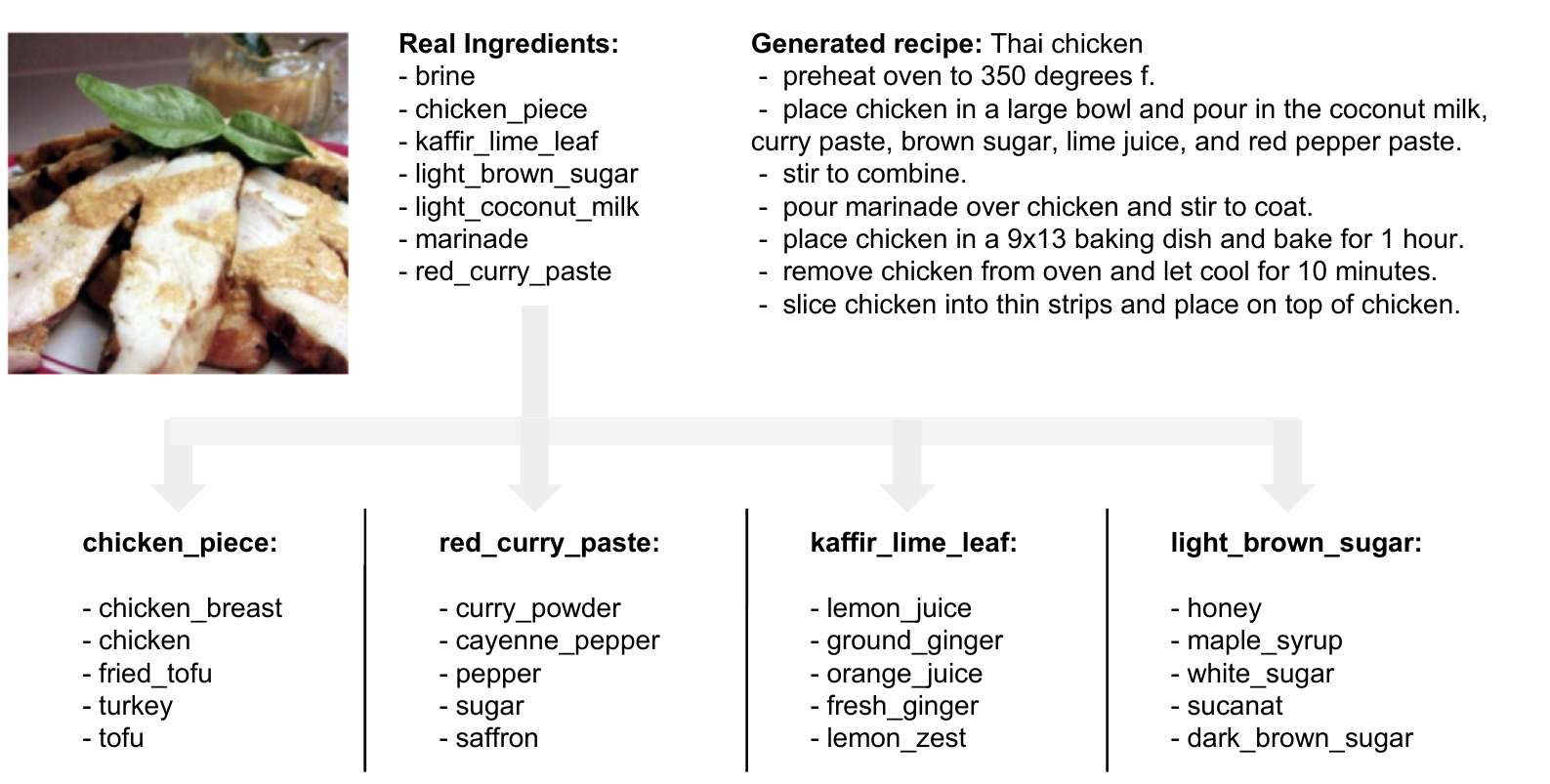}
   \caption{%
   \label{fig:top_5_ground_truth} %
    Top 5 suggestions of \GIN{} for 4 different ingredients on a Thai chicken recipe. Here, \GIN{} is used on the ground truth ingredients of a recipe. \GIN{}'s suggestions to replace \emph{chicken} include vegetarian options although the \emph{chicken piece} ingredient itself never appears together with \emph{tofu} in the training set.}
\end{figure*}

\begin{figure*}[b]
   \centering
   \includegraphics[width=0.98\textwidth]{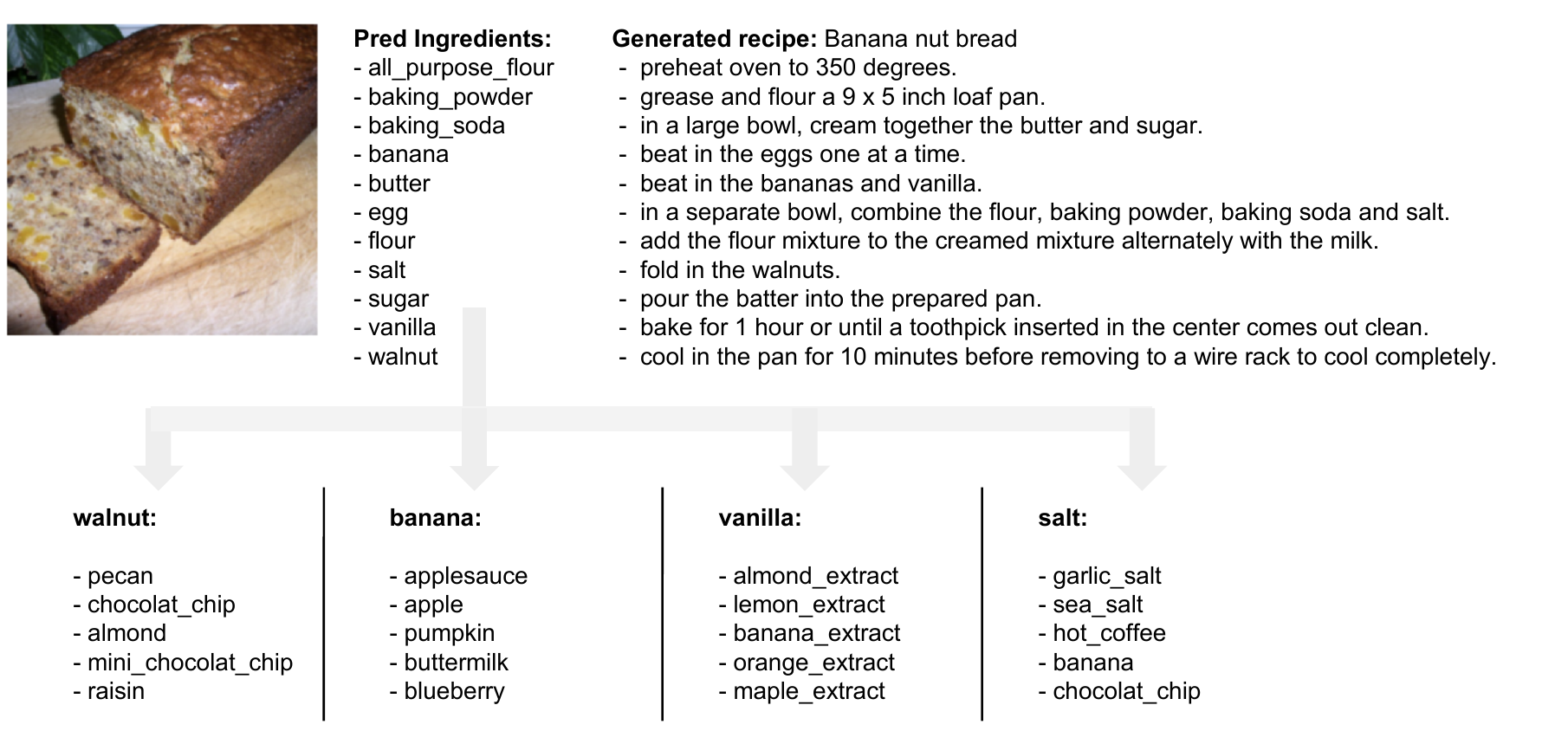}
   \caption{%
   \label{fig:top_5_pred_ingredients} %
    Top 5 suggestions of \GIN{} for 4 different ingredient substitutions. In this case, \GIN{} takes the predicted ingredients of the inverse cooking pipeline as input. \GIN{}'s replacement suggestions for \emph{walnuts} include other types of \emph{nuts} as well as \emph{chocolate chips} and \emph{raisin}, which could be good replacements in this context.}
\end{figure*}

\begin{figure*}[b]
   \centering
   \includegraphics[width=0.98\textwidth]{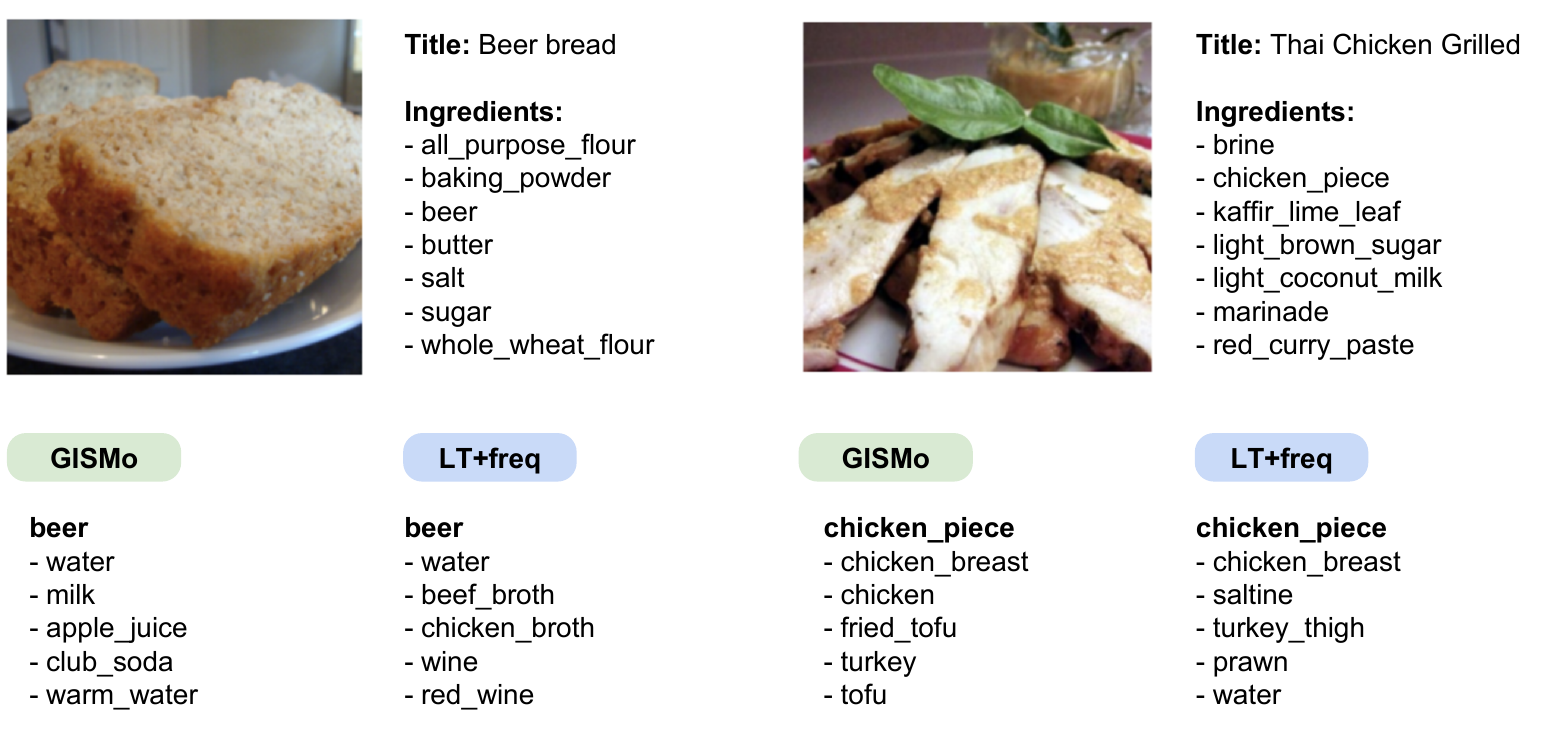}
   \caption{%
   \label{fig:lt_freq_vs_gismo_id} %
    Qualitative comparison of the performance of \GIN{} against the LT-Freq approach by showing the top-5 suggestions for each approach on two examples where the ingredient to substitute (\emph{beer} or \emph{chicken piece}) appears in the training set (in distribution). \GIN{} can provide better substitutions by leveraging the context: \emph{chicken broth}, although often associated with \emph{water} or \emph{beer}, is not a good substitution in the context of a cake.}
\end{figure*}

\begin{figure*}[b]
   \centering
   \includegraphics[width=0.98\textwidth]{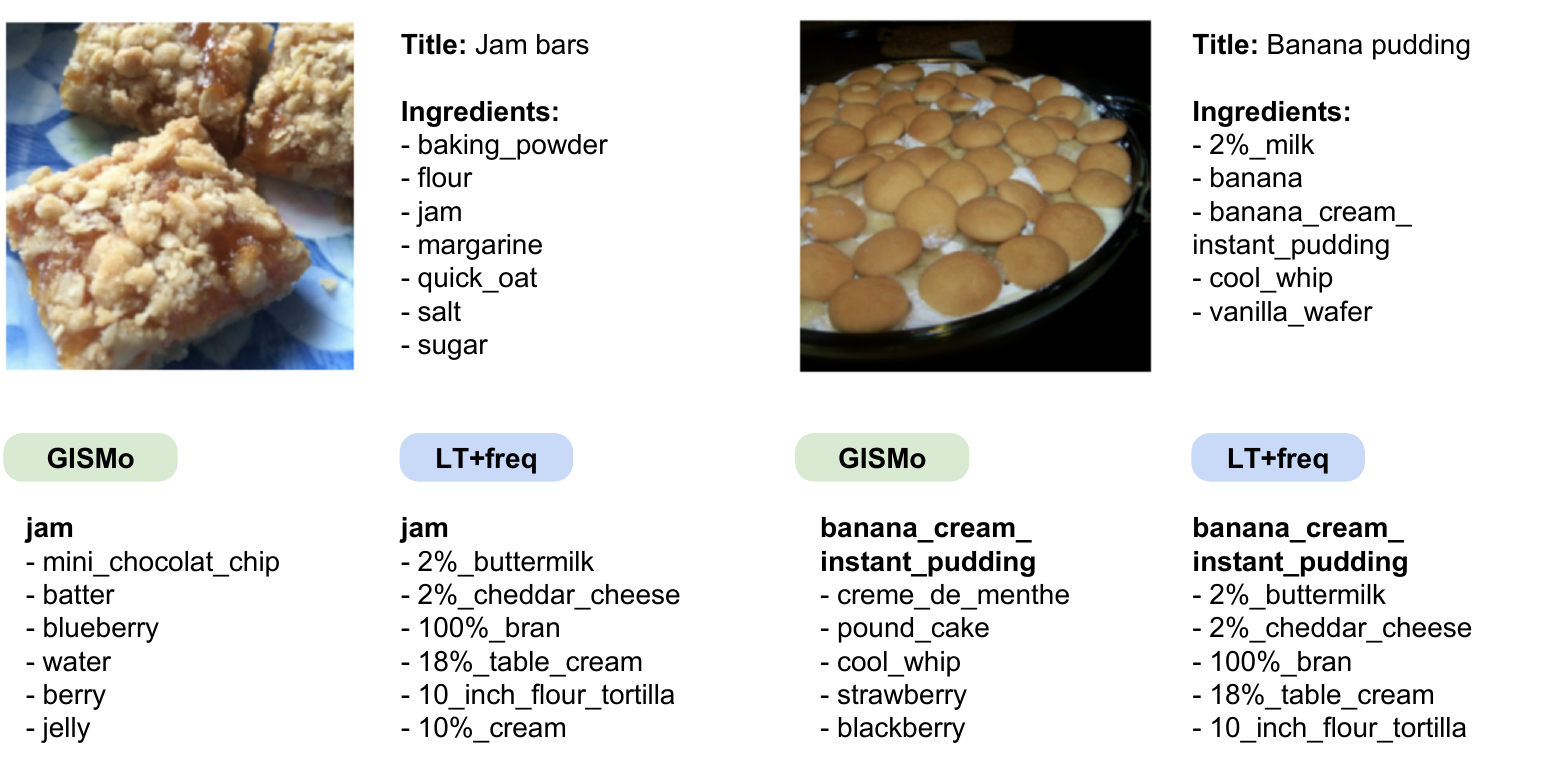}
   \caption{%
   \label{fig:lt_freq_vs_gismo_ood} %
    Comparing the performance of \GIN{} against the LT-freq approach by showing the top-5 suggestions for each approach on examples where the ingredient to substitute (\emph{jam} or \emph{banana cream instant pudding}) does not appear in the training set (out of distribution). \GIN{} is able to provide creative substitutions for ingredients never seen at the training time, whereas the LT-freq approach will fall in a default mode (returning ingredients by alphabetical order). Note that approximately half of the ingredients in the ingredients vocabulary never appear in the training set, so this generalization to unseen ingredients is very important.}
\end{figure*}

We first look at the ability of \GIN{} to provide interesting and plausible substitutions. Figure \ref{fig:top_5_ground_truth} shows substitution examples when using ground truth ingredients as input, and Figure \ref{fig:top_5_pred_ingredients} depicts the predicted substitutions when using ingredients predicted by the personalized inverse cooking pipeline as input. In both cases, \GIN{} comes up with reasonable substitutions given the recipe context. We also compare the suggestions of \GIN{} to those of the LT-Freq (Lookup Table + Frequency) approach, our strongest baseline, in Figures \ref{fig:lt_freq_vs_gismo_id} -- \ref{fig:lt_freq_vs_gismo_ood}, showing qualitatively that \GIN{} can adapt to context and to propose substitutions not seen during training.

\begin{figure*}[b]
   \centering
   \includegraphics[width=0.98\textwidth]{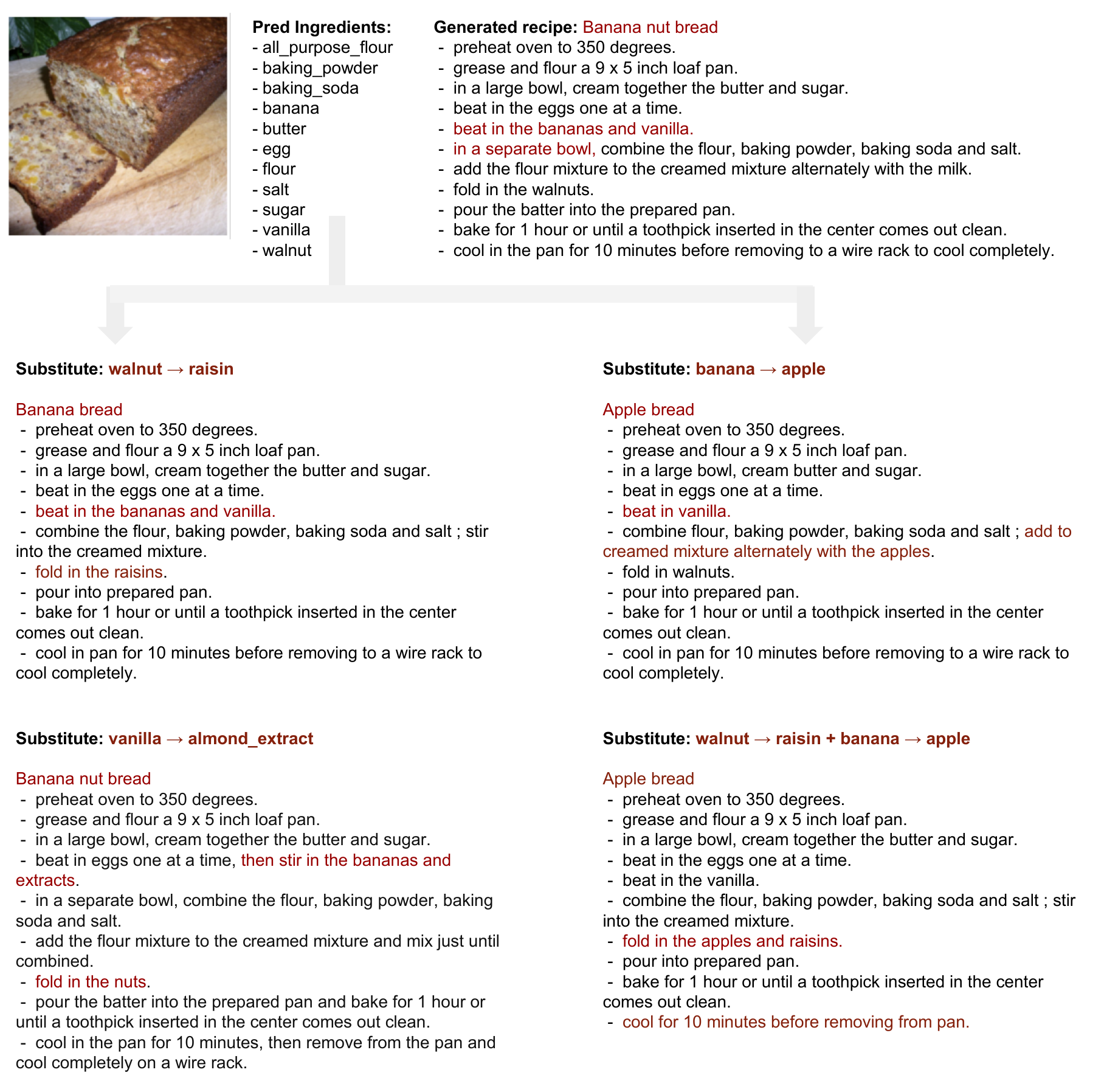}
   \caption{%
   \label{fig:trying_top_5_suggestions} %
    Trying out some of the replacement suggestions of \GIN{} for \emph{walnut} and \emph{banana} in a ``banana nut bread recipe''. In the bottom right example, we try to combine two \GIN{} substitutions suggestions. The title is correctly adapted in all cases, and instructions changes go beyond searching and replacing ingredients.}
\end{figure*}

\begin{figure*}[b]
   \centering
   \includegraphics[width=0.98\textwidth]{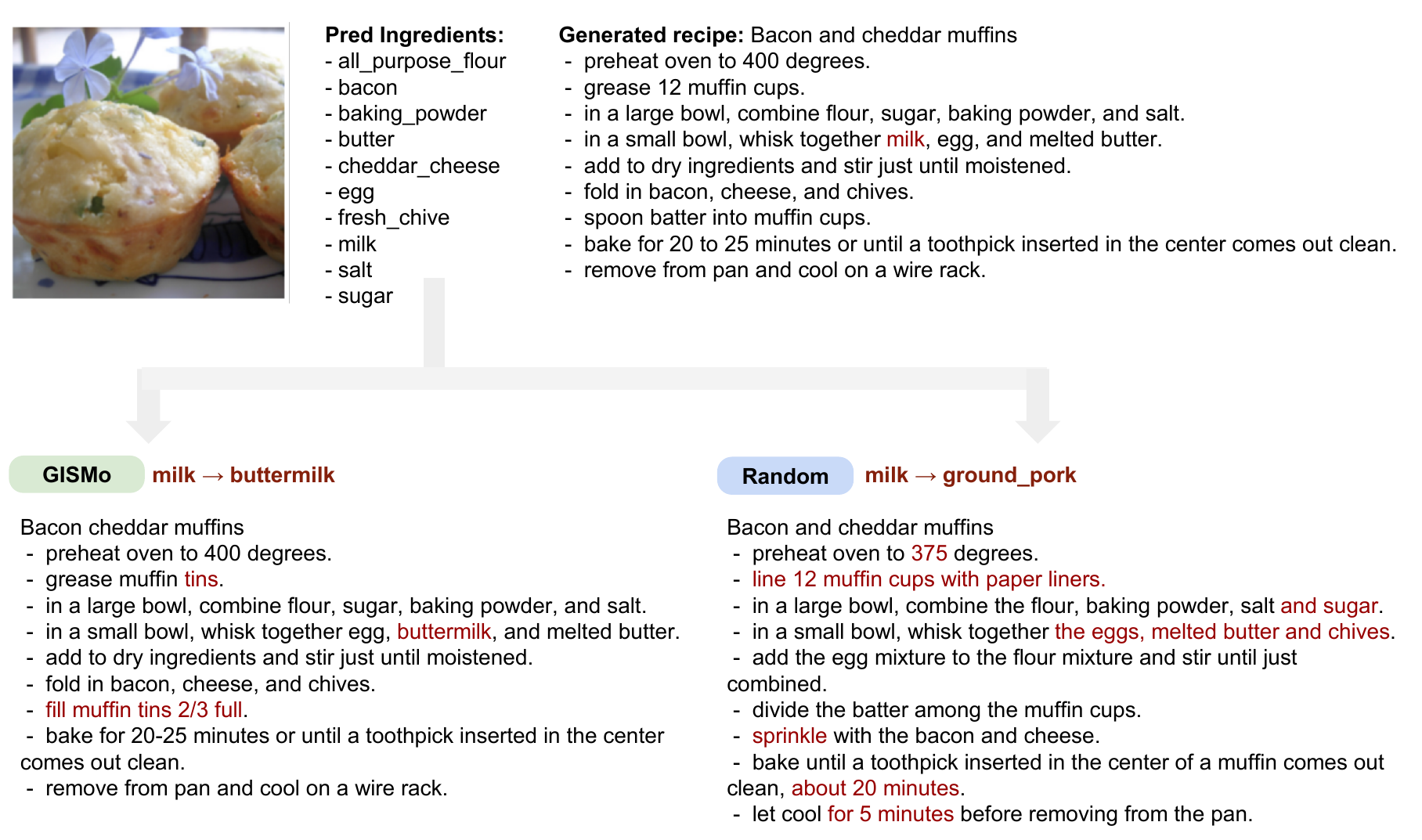}
   \caption{%
   \label{fig:random_subsitutions} %
    Comparison of the effect of providing correct substitutions vs incorrect substitutions on the recipe generation process. Substituting \emph{beer} with a random ingredient (here \emph{ground pork}) instead of relying on \GIN{}, leads to the correct removal of the substituted ingredient, but the new ingredient is ignored. This is in line with our metrics in which we saw that random substitutions decrease the IoU between ingredients used to generate the recipe and ingredients appearing in the generated recipe.}
\end{figure*}

We then select some of the top-5 substitutions proposed by \GIN{} in Figure \ref{fig:top_5_pred_ingredients} to generate new recipes with the personalized inverse cooking pipeline, and experiment with the effect of replacing two ingredients with \GIN{}'s suggestions instead of just one. The results are shown in Figure \ref{fig:trying_top_5_suggestions}. We extend this qualitative experiment to highlight the importance of using reliable candidate substitutions from \GIN{}, as opposed to random substitutions. Figure \ref{fig:random_subsitutions} presents these results. Note that random substitutions are often ignored by the recipe generator, leading to lower IoU (consistency) scores.\looseness-1

\begin{figure*}[b]
   \centering
   \includegraphics[width=0.98\textwidth]{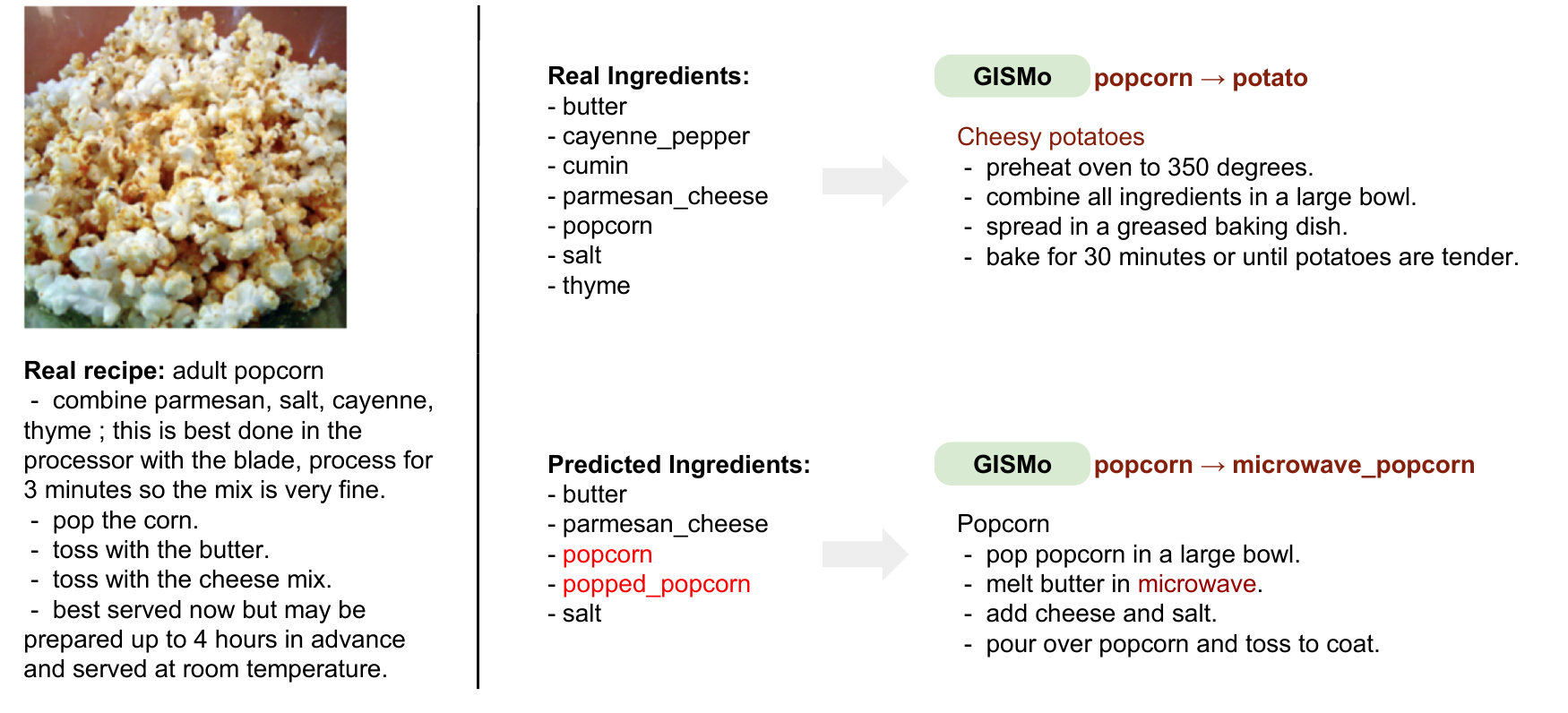}
   \caption{%
   \label{fig:duplicate_popcorn} %
    \GIN{} can leverage the context to come up with creative ingredient substitutions. Here we show the top-1 suggestion of \GIN{} for replacing \emph{popcorn} in a ``popcorn recipe''. \GIN{} suggests \emph{potato} as substitution, and the personalized inverse cooking pipeline creates a ``cheesy potato recipe''. The ability of \GIN{} to come up with interesting substitutions may be negatively impacted by the incomplete, wrong or near duplicate ingredient predictions (second row) resulting from the inverse cooking pipeline, as \GIN{} leverages these predictions as context.}
\end{figure*}

\begin{figure*}[b]
   \centering
   \includegraphics[width=0.98\textwidth]{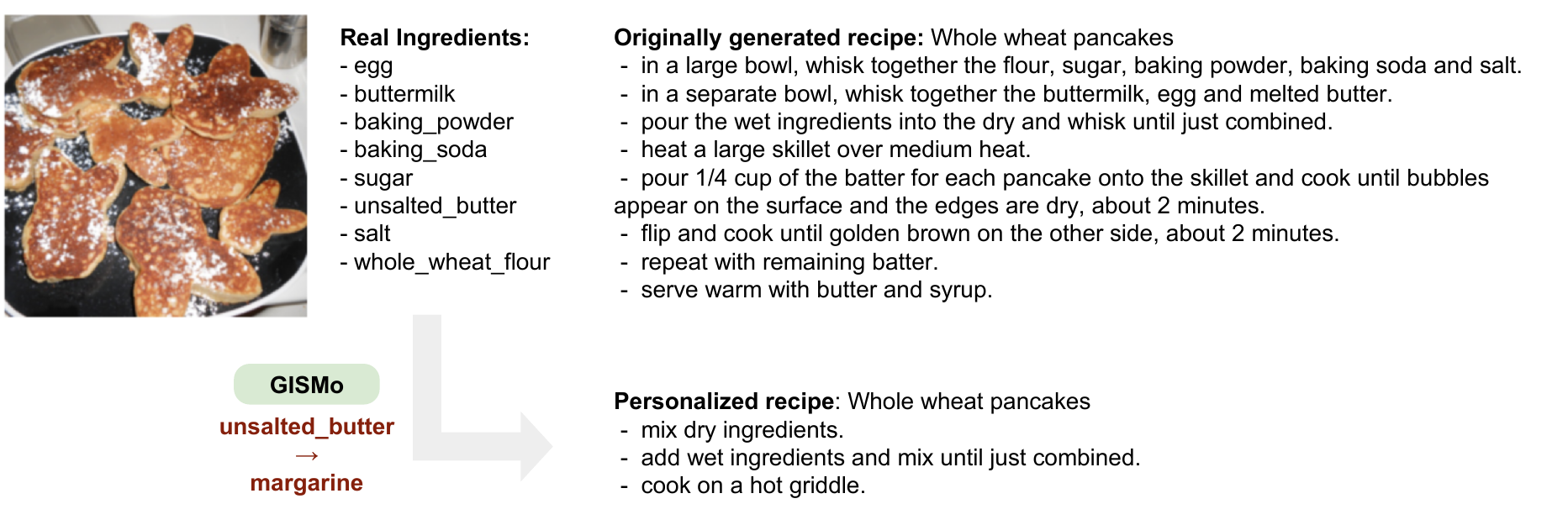}
   \caption{%
   \label{fig:collapse_on_substitutions} %
    Example of recipe collapse when performing a substitution that would appear to be minor. Although the title is preserved, the instructions of the personalized recipe are generic, lacking all the details of the originally generated recipe.}
\end{figure*}

\begin{figure*}[b]
   \centering
   \includegraphics[width=0.98\textwidth]{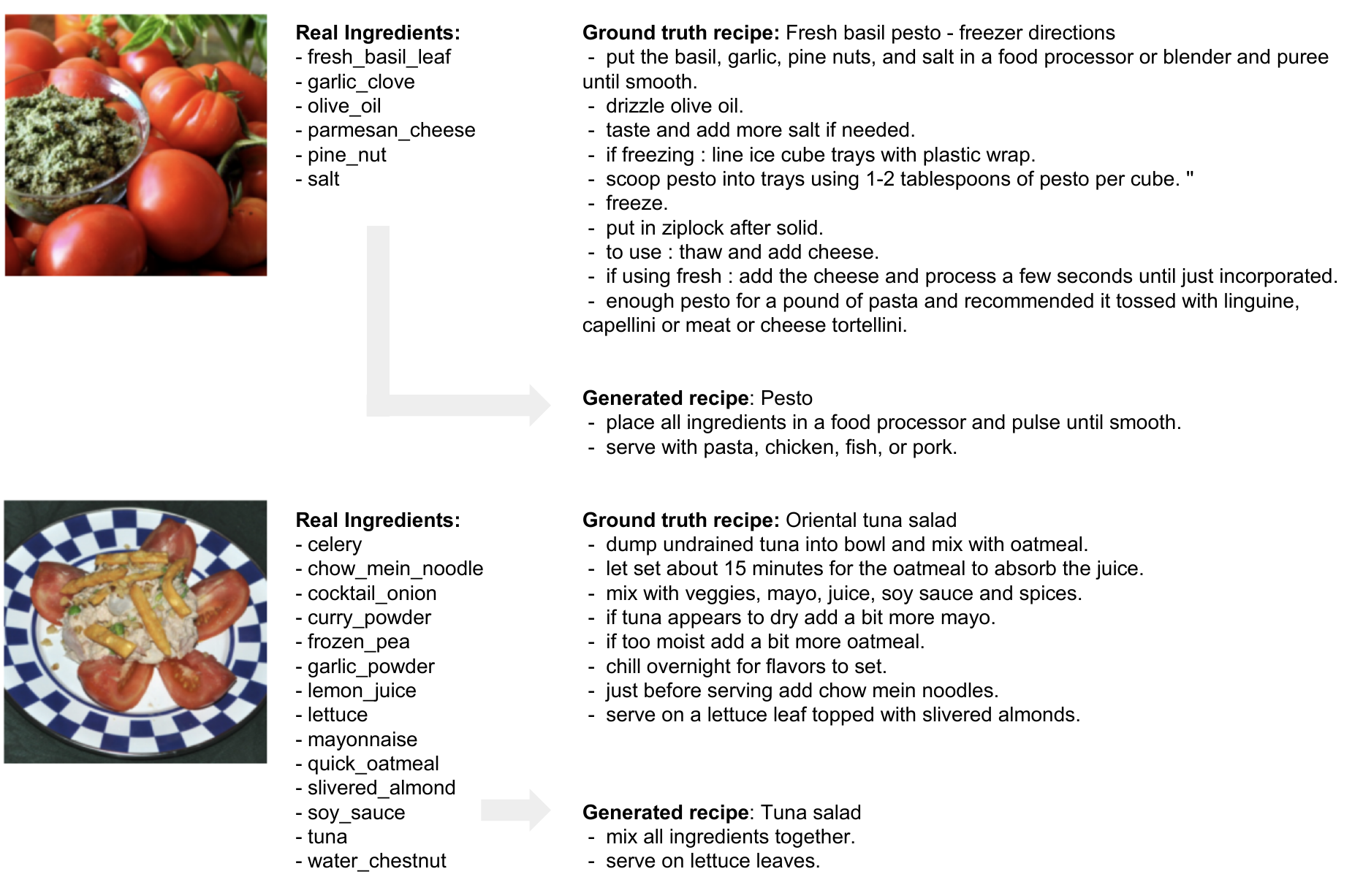}
   \caption{%
   \label{fig:collapse_without_subs_figure} %
    The recipe collapse is not necessarily due to the proposed ingredient substitutions. Here we show examples of recipe collapse occurring when leveraging ground truth ingredients without substitutions to condition the recipe generation process.}
\end{figure*}

Finally, we look at some failure modes of the personalized inverse cooking pipeline when used in combination with \GIN{}, pointing to potential areas of improvement. Figure \ref{fig:duplicate_popcorn} shows how the creativity of \GIN{} to come up with ingredient substitutions might be negatively impacted by incomplete, wrong, or near duplicate ingredient predictions of the personalized inverse cooking pipeline. Recall that, in this case, the ingredients predicted by the inverse cooking pipeline constitute the context of \GIN{}, therefore impacting the kind of substitutions that \GIN{} can propose. Improving the ingredient prediction step in the personalized inverse cooking pipeline holds the promise to improve \GIN{}'s suggestions as a result. Moreover, Figure \ref{fig:collapse_on_substitutions} shows the phenomenon of recipe collapse that sometimes occurs when performing (even minor) substitutions with the personalized inverse cooking pipeline. However, as shown in Figure \ref{fig:collapse_without_subs_figure}, this phenomenon also appears in the inverse cooking pipeline which does not consider any substitution, pointing to potential areas of improvement of the recipe generator module of the personalized inverse cooking pipeline.\looseness-1

\end{document}